\documentclass[conference]{IEEEtran}
\IEEEoverridecommandlockouts
\usepackage{cite}
\usepackage{amsmath,amssymb,amsfonts}
\usepackage{algorithmic}
\usepackage{threeparttable}
\usepackage{graphicx}
\usepackage{textcomp}
\usepackage{multirow, multicol}
\usepackage{booktabs} 
\usepackage{xcolor}
\usepackage{cite}
\usepackage{verbatim}
\usepackage{subfigure}

\def\BibTeX{{\rm B\kern-.05em{\sc i\kern-.025em b}\kern-.08em
    T\kern-.1667em\lower.7ex\hbox{E}\kern-.125emX}}
    
\begin{document}

\title{3D-LSPTM: An Automatic Framework with 3D-Large-Scale Pretrained Model for Laryngeal Cancer Detection Using Laryngoscopic Videos

\thanks{This work is partially supported by the Basic and Applied Basic Research Project of Guangdong Province (2022B1515130009), the Special subject on Agriculture and Social Development, Key Research and Development Plan in Guangzhou (2023B03J0172), the Natural Science Foundation of Top Talent of SZTU (GDRC202318), the Research Promotion Project of Key Construction Discipline in Guangdong Province (2022ZDJS112), and the Key Topics of the 13th Five Year Plan for National Education Science in 2019(DHA190440).}
\thanks{Meiyu Qiu, Wenjun Huang, Haoyun Zhang, Xiaomao Fan are with the College of Big Data and Internet, Shenzhen Technology University, Shenzhen, China (e-mail: 201902010408@stumail.sztu.edu.cn; 202100202165@stumail.sztu.edu.cn; 202100202164@stumail.sztu.edu.cn; fanxiaomao@sztu.edu.cn).}
\thanks{Yun Li and Wenbin Lei are with the First Affiliated Hospital,Sun Yat-Sen University, Guangzhou, China (e-mail:liyun76@mail.sysu.edu.cn; leiwb@mail.sysu.edu.cn).}
\thanks{Weiping Zheng is with the School of Computer Science, South China Normal University, Guangzhou, China (email:zhengweiping@scnu.edu.cn).}
\thanks{Meiyu Qiu is also with the School of Applied Technology, Shenzhen University, Shenzhen, China.}
\thanks{Meiyu Qiu and Yun Li contribute equally.}
\thanks{Xiaomao Fan and Wenbin Lei are the co-corresponding authors.}}

\author{\IEEEauthorblockN{Meiyu Qiu, Student member, IEEE}
\and
\IEEEauthorblockN{Yun Li}
\and
\IEEEauthorblockN{Wenjun Huang}
\and
\IEEEauthorblockN{Haoyun Zhang}
\and
\IEEEauthorblockN{Weiping Zheng}
\and
\IEEEauthorblockN{Xiaomao Fan, Member, IEEE}
\and
\IEEEauthorblockN{Wenbin Lei}
}

\maketitle

\begin{abstract}
Laryngeal cancer is a malignant disease with a high morality rate in otorhinolaryngology, posing an significant threat to human health. Traditionally larygologists manually visual-inspect laryngeal cancer in laryngoscopic videos, which is quite time-consuming and subjective. In this study, we propose a novel automatic framework via 3D-large-scale pretrained models termed 3D-LSPTM for laryngeal cancer detection. Firstly, we collect 1,109 laryngoscopic videos from the First Affiliated Hospital Sun Yat-sen University with the approval of the Ethics Committee. Then we utilize the 3D-large-scale pretrained models of C3D, TimeSformer, and Video-Swin-Transformer, with the merit of advanced featuring videos, for laryngeal cancer detection with fine-tuning techniques. Extensive experiments show that our proposed 3D-LSPTM can achieve promising performance on the task of laryngeal cancer detection. Particularly, 3D-LSPTM with the backbone of Video-Swin-Transformer can achieve 92.4\% accuracy, 95.6\% sensitivity, 94.1\% precision, and 94.8\% $F_1$. 
\end{abstract}


\section{Introduction}
Laryngeal cancer is a malignant disease in otorhinolaryngology, which is associated with bad living habits. Researches show that the estimated number of new cases is 12,570 in 2022 with the estimated number of death is 3,820\cite{divakar2023trends}, which presents a high risk of mortality, posing an significant threat to human health. In clinical, laryngoscopy biopsy serves as a golden standard for laryngeal cancer detection\cite{steuer2017update}. In recent years, advancements in medical technology have led to the development of laryngendoscopes, like stroboscopic laryngendoscope, which have proven to be effective tools for laryngeal cancer detection. Laryngologists manually detect the laryngeal cancer through the recorded laryngoscopic videos, which is time-consuming and it mainly depends on the experience and expertise of the laryngologists. These challenges are particularly pronounced in remote areas where are short of medical sources and experienced laryngologists, it is inevitable to lead to missed diagnosis and misdiagnosis. Therefore, developing an automatic laryngeal cancer detection model is of great necessarity. 

With the rapid advancement of deep learning techniques, researchers have increasingly turned to novel approaches based on deep learning for the analysis of medical images and videos, including the detection of laryngeal cancer. Traditional methods for computer vision (CV) tasks primarily rely on Convolutional Neural Networks (CNNs) \cite{ronneberger2015u,tran2015c3d, cciccek20163d-unet,mehta2018net, huang2020unet,valanarasu2020kiu}, which have demonstrated remarkable success in vision tasks. However, CNN-based methods suffer from limitations in capturing long-range dependencies beyond the receptive field. They are only capable of capturing dependencies within the receptive field and often overlook crucial contextual information.

The emergence of the Transformer \cite{vaswani2017attention} has addressed this limitation to some extent, introducing the attention mechanism and achieving impressive performance in computer vision tasks compared to CNNs. As a result, researchers have modified the Transformer architecture to adapt it to various vision tasks \cite{liu2021swin,touvron2021training,wu2021rethinking,valanarasu2021medical, chen2021transunet}, achieving competitive results. Medical-Transformer\cite{valanarasu2021medical} propose a position-sensitive axial attention mechanism and a local-global training strategy, which can achieve a high medical images segmentation performance on small medical datasets. TransUNet\cite{chen2021transunet} combines Transformer and U-Net through utilizing the information of the image features and CNN features, which can also achieves competitive medical image segmentation results. However, Transformer is not well-suited for certain spatial-temporal video understanding tasks, such as video classification, due to its low-resolution feature maps. The success of transformer-based methods in image classification has prompted further exploration in video recognition tasks \cite{neimark2021video, bertasius2021times, arnab2021vivit, fan2021multiscale,zhang2021multi,touvron2021deiT}. These approaches mostly build upon the Transformer backbone and incorporate spatial attention mechanisms for spatial-temporal understanding. For instance, VTN \cite{neimark2021video} introduced a temporal attention encoder block that achieved high performance in video action recognition tasks. Timesformer \cite{bertasius2021times} employed a divided spatial-temporal self-attention mechanism based on Transformers, reducing training costs while achieving efficient results. ViViT \cite{arnab2021vivit} proposed several variants of Transformer that factorize the spatial and temporal dimensions of the input, leading to improved performance and computational efficiency. To address the challenges of high-resolution inputs in Transformer, MViT \cite{fan2021multiscale} utilized pooling attention techniques to aggregate local features. These algorithms primarily focus on action recognition tasks and have demonstrated promising results. However, one common drawback among these approaches is the use of global self-attention modules, resulting in significant computation and memory overhead. To tackle this issue, Video-Swin-Transformer \cite{liu2022videoswin} adopts an effective shifted window mechanism and locality bias from \cite{liu2021swin}, achieving excellent performance in action recognition tasks.

In this study, we first collect 1,109 laryngoscopic videos from the First Affiliated Hospital Sun Yat-sen University (FAHSYU) with the approval of the Ethics Committee of FAHSYU (approval no. [2021]416). Then we develop an advanced automatic framework termed 3D-LSPTM for the detection of laryngeal cancer, utilizing 3D-large-scale pretrained models. The 3D-LSPM incorporates three state-of-the-art video recognition algorithms of C3D, TimeSformer, and Video-Swin-Transformer, with the merit of advanced featuring videos, to successfully accomplish the classification of medical videos. The experimental results have demonstrated that the Video-Swin-Transformer model exhibits exceptional performance in detecting laryngeal cancer, offering significant potential to assist laryngologists in promptly and accurately identifying this condition.

To sum up, the main contributions of this study can be summarized as follows:

\begin{itemize}
    \item We propose a novel automatic approach via 3D-large-scale pretrained models called 3D-LSPTM for laryngeal cancer detection, which can well capture the contextual information of laryngoscopic videos with the advantages of advanced featuring videos.

    \item We have meticulously curated a real dataset of laryngoscopic videos comprising 1,109 samples, which are collected from the First Affiliated Hospital Sun Yat-sen University. To ensure the highest quality, each video in the dataset has been carefully labeled, providing valuable annotations for further analysis.
    
    \item Through extensive experiments conducted on our meticulously collected laryngoscopic video dataset, we have demonstrated that the 3D-LSPTM consistently achieves competitive performance. Notably, when combined with the Video-Swin-Transformer backbone, the 3D-LSPTM surpasses other models in terms of performance metrics.
\end{itemize}

\section{Methods}
\begin{figure}
\centering
\includegraphics[width=\linewidth]{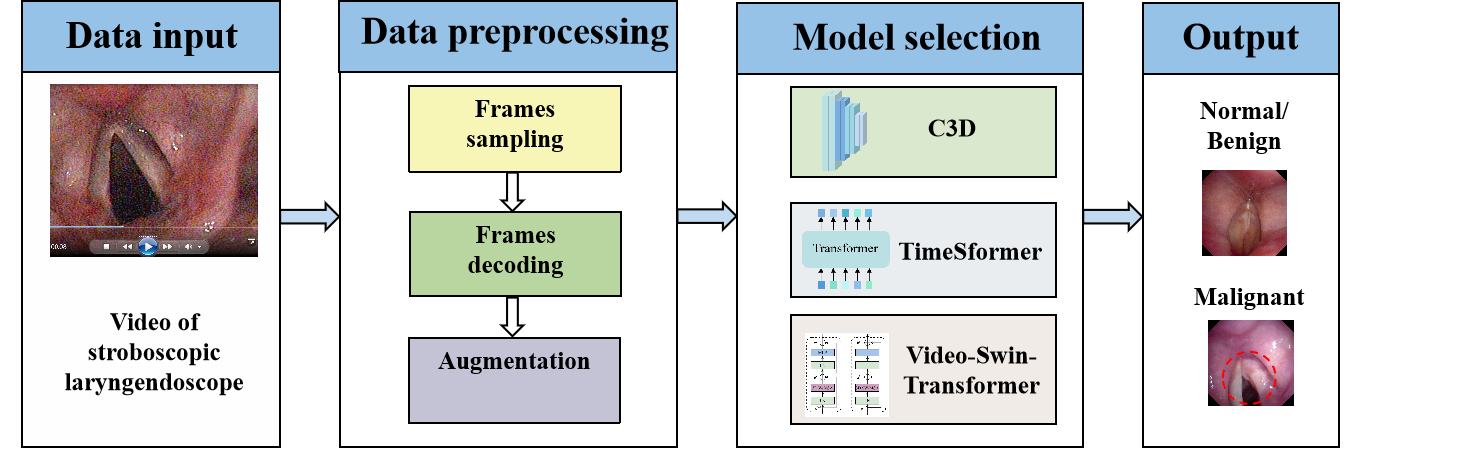}
\caption{The pipeline of 3D-LSPTM.}
\label{framework}
\end{figure}

Fig. \ref{framework} shows the overall pipeline of the proposed 3D-LSPTM. It mainly consists of data acquisition, data preprocessing, and model selection, which are described in detail as following:

\subsection{Data acquisition}
This study was approved by the Ethics Committee of FAHSYU (approval no. [2021]416). All laryngoscopic videos in this dataset were obtained from consecutive patients aged $\ge$ 18 years who underwent fiberoptic laryngoscopy in FAHSYU and stored in AVI format. This dataset contained 1,109 videos, which was consisted of 555 normal cases, 240 benign cases, and 314 malignant cases. The judgement was made according to the pathologic reports by highly experienced laryngoscopists, who had at least five years of experience in laryngoscopy and performing more than 3,000 laryngoscopy examinations. More important, the pathological diagnosis was confirmed by two board-certified pathologists using haematoxylin-eosin-stained tissue slides according to the World Health Organization classification of tumors, which served as the gold standard for judgment. The detection of laryngeal cancer is crucial, so we divided the dataset into laryngeal cancer (\emph{i.e.}, malignant cases) and non-laryngeal cancer (\emph{i.e.}, normal and benign cases), represented by 0 or 1, to better determine the accuracy of the model in identifying laryngeal cancer.

\subsection{Data preprocessing}
Since videos are compose of a series of continuous frames, processing video inputs as images can be more effective. Therefore, we firstly sample frames by spliting the laryngoscopic videos into frames. The numbers of the frames are determined by the length of the videos. For each video, we select 32 frames with the frame interval of 2 as the input for our model. And these frames are subsequently decoded into raw image data. Finally we apply images flipping and resize the frames to a scale of 224 $\times$ 224 pixels. 

\subsection{Model selection}
Extensive evidence \cite{fan2021multiscale} has unequivocally demonstrated the superiority of pre-trained models over models trained from scratch. Building upon this knowledge, our study introduces the 3D-LSPTM, which leverages the power of three widely acclaimed large-scale pretrained models: C3D, TimeSformer, and Video-Swin-Transformer. Below, we provide a brief overview of each of these models:

\subsubsection{C3D}
C3D \cite{tran2015c3d} is a pioneering work which utilize the 3D-convolution layers to model the temporal feature of the video. C3D employ a spatial and temporal feature learning technique, which can achieve a good performance on kinds of video analysis tasks like action recognition, scene and object recoginition. And extensive experiments have shown that the kernel size of 3 $\times$ 3 of the receptive field yields the best results. For the laryngeal cancer detection problem, it is a binary classification to detect the patients with laryngencancer or not.

\subsubsection{TimeSformer}
TimeSformer \cite{bertasius2021times} followed the scheme of Transformer, viewing the video as a sequence of continuous frame-level patches, which are feeding as the input token embeddings to the Transformer. To verify the performance of the attention block on video analysis tasks, TimeSformer proposed five different space-time self-attention designs, specifically space attention, joint space-time attention, divided space-time attention, sparse local global attention, and axial attention. Experiment results show that the divided space-time attention architecture achieves the best performance among the five attention blocks. In this study, we also utilize this paradigm to train our laryngoscopic video dataset.

\subsubsection{Video-Swin-Transformer}
Multi-head self-attention (MSA) mechanisms on each non-overlapping 2D window has been shown to be both effective and efficient for image recognition, so Video-Swin-Transformer\cite{liu2022videoswin} extend this design for video input task. However, this mechanism lacks of the connections of the different windows and it can only catch the local information. To solve this problem, Video-Swin-Transformer extends the 2D shifted windows mechanism of Swin Transformer\cite{liu2021swin} to the 3D shifted windows. Meanwhile, since the original global attention block for video classification tasks will cause large computation and memory cost, Video-Swin-Transformer follows the paradigm of Swin Transformer to introduce a inductive bias of locality, which is shown effective in the video tasks.

\section{Experiment results and discussion}

\subsection{Computing environment}
All the experiments were executed on a high performance computing server equipped with Intel(R) Xeon Silver 4116R CPU @ 2.10GHz, 125GB host memory and one piece of NVIDIA GeForce RTX 3090 graphics processing unit (GPU) card with 24GB device memory. The computing server installed a Ubuntu system at version 20.04.6, with Python 3.8.16 and CUDA 11.2, as well as the prevalent deep learning framework Pytorch 2.11.1. 

\subsection{Classification performance}
In this study, we employed a rigorous ten-fold cross-validation approach on our laryngoscopic video dataset to ensure robust evaluation. To assess the classification performance, we utilized four widely recognized metrics: accuracy ($Acc$), sensitivity ($Sen$), precision ($Pre$), and the $F_1$ score.

To train the 3D-LSPTM model, we leveraged the knowledge encoded in three large-scale pretrained models: C3D, TimeSformer, and Video-Swin-Transformer. The experimental results, illustrated in TABLE \ref{tab:comparison}, highlight the performance of each model. C3D achieved notable scores across the metrics, with accuracy, sensitivity, precision, and $F_1$ values of 87.4\%, 94.2\%, 89.1\%, and 91.5\%, respectively. The classification performance of TimeSformer demonstrated improvement over C3D, surpassing it by a margin of 1.1\% in accuracy, 1.3\% in sensitivity, 0.5\% in precision, and 0.9\% in $F_1$ score. Notably, Video-Swin-Transformer exhibited exceptional performance, achieving an impressive accuracy of 92.4\%, sensitivity of 95.6\%, precision of 92.1\%, and $F_1$ score of 94.8\%. These results clearly establish Video-Swin-Transformer as the superior method, outperforming both C3D and TimeSformer, with a substantial margin of 5.0\% in accuracy compared to C3D and 3.9\% in accuracy compared to TimeSformer.

\begin{figure*}[tb]
  \centering
  \subfigure[]{
   \includegraphics[width=0.31\linewidth]{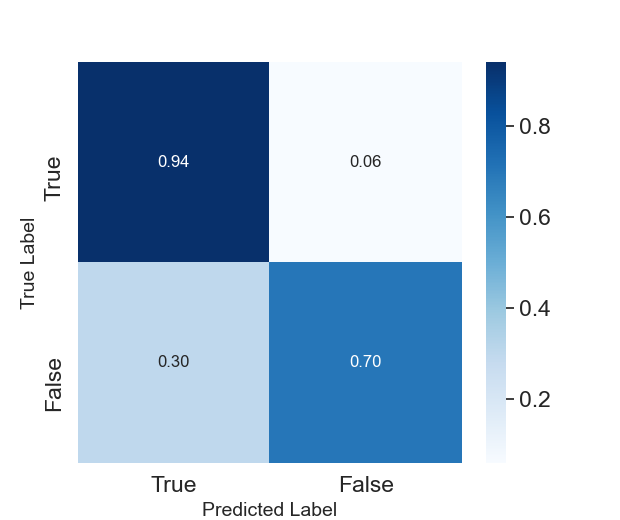}
  }
  \subfigure[]{
   \includegraphics[width=0.31\linewidth]{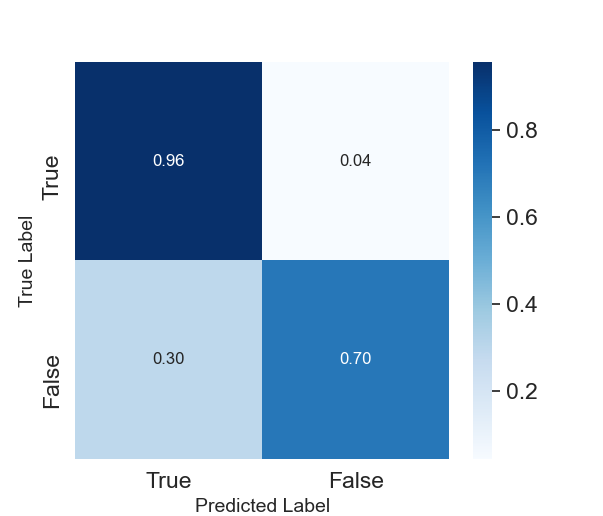}
  }
   \subfigure[]{
   \includegraphics[width=0.31\linewidth]{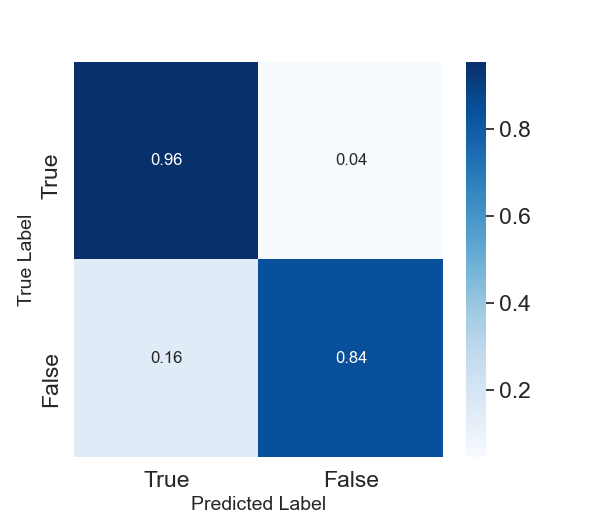}
  }
  \caption{Confusion matrix of 3D-LSPTM with three backbones: (a) C3D, (b) TimeSformer, and (3)Video-Swin-Transformer.}
   \label{matrix}
\end{figure*}

Furthermore, Fig. \ref{matrix} presents the confusion matrices of the three models, providing additional insights into their performance. It is evident that both TimeSformer and Video-Swin-Transformer yield superior results compared to C3D. This observation aligns with the fact that TimeSformer and Video-Swin-Transformer, being Transformer-based models, possess stronger representation abilities than C3D, a CNN-based model, owing to the attention mechanism inherent in Transformers. Notably, the classification of normal or benign cases achieved better results than that of malignant cases, with differences of 24\%, 16\%, and 12\% for C3D, TimeSformer, and Video-Swin-Transformer, respectively. This further emphasizes the excellent classification capability of Video-Swin-Transformer, which holds great potential for medical video classification tasks in future research endeavors.

\begin{table}[bt]
    \centering
    \caption{Classification Performance comparison for laryngeal cancer detection.}
    \renewcommand{\arraystretch}{1.5}
    \begin{tabular}{ccccc}
    \toprule 
    Model & $Acc$ & $Sen$ & $Pre$ & $F_1$\\
    \midrule 
    C3D\cite{tran2015c3d}& 0.874 &0.942 &0.891 &0.915\\
    TimeSformer\cite{bertasius2021times}&0.885 & 0.957 &0.896 &0.924\\
    Video-Swin-Transformer\cite{liu2022videoswin}&0.924 &0.956 &0.941 &0.948\\
    \bottomrule 
    \end{tabular}
    \label{tab:comparison}
\end{table}

\section{Conclusion}
In this study, we present an innovative and automated framework termed 3D-LSPTM for the detection of laryngeal cancer using laryngoscopic videos. Our framework leverages the power of three state-of-the-art pretrained models: C3D, TimeSformer, and Video-Swin-Transformer. Notably, this is the first time where action recognition algorithms have been employed to accomplish the challenging task of medical video classification in the context of laryngoscopic videos. Through extensive experimentation and evaluation, we have obtained compelling results that demonstrate the adaptability and effectiveness of our proposed framework for automatic laryngeal cancer detection. In particular, the 3D-LSPTM integrating Video-Swin-Transformer has yielded remarkable outcomes. These findings hold significant promise for both laryngologists and patients, as it paves the way for improved and efficient laryngeal cancer detection.

\end{document}